\documentclass{article}

\usepackage{microtype}
\usepackage{graphicx}
\usepackage{tikz}
\usepackage{subfigure}
\usepackage{booktabs} %

\usepackage{hyperref}

\usepackage[accepted]{icml2024}%

\usepackage{amsmath,amsfonts,bm}

\def\Figref#1{Figure~\ref{#1}}

\def\secref#1{section~\ref{#1}}

\def\eqref#1{equation~\ref{#1}}

\def\1{\bm{1}}

\DeclareMathAlphabet{\mathsfit}{\encodingdefault}{\sfdefault}{m}{sl}
\SetMathAlphabet{\mathsfit}{bold}{\encodingdefault}{\sfdefault}{bx}{n}

\def\gA{{\mathcal{A}}}

\def\gN{{\mathcal{N}}}

\def\gS{{\mathcal{S}}}

\def\gZ{{\mathcal{Z}}}

\DeclareMathOperator*{\argmax}{arg\,max}

\usepackage{graphicx}

\newcommand{\mysubsubsection}[1]{{\bf #1}\ }

\renewcommand{\printAffiliationsAndNotice}[1]{%
{\let\thefootnote\relax\footnotetext{#1}
}
}

\icmltitlerunning{Large Language Models as Agents in Two-Player Games}

\begin{document}

\twocolumn[
\icmltitle{Large Language Models as Agents in Two-Player Games}

\begin{icmlauthorlist}
\icmlauthor{Yang Liu$^*$ }{}
\icmlauthor{Peng Sun$^*$}{}
\icmlauthor{Hang Li}{}
\end{icmlauthorlist}
\begin{center}
~\\
\textbf{ByteDance Research}  \\
\{yang.liu01,~wanhesong,~lihang.lh\}@bytedance.com
\end{center}

\vskip 0.3in
]
\printAffiliationsAndNotice{\icmlEqualContribution}

\begin{abstract}
By formally defining the training processes of large language models (LLMs), which usually encompasses pre-training, supervised fine-tuning, and reinforcement learning with human feedback, within a single and unified machine learning paradigm, we can glean pivotal insights for advancing LLM technologies. This position paper delineates the parallels between the training methods of LLMs and the strategies employed for the development of agents in two-player games, as studied in game theory, reinforcement learning, and multi-agent systems. We propose a re-conceptualization of LLM learning processes in terms of agent learning in language-based games. This framework unveils innovative perspectives on the successes and challenges in LLM development, offering a fresh understanding of addressing alignment issues among other strategic considerations. Furthermore, our two-player game approach sheds light on novel data preparation and machine learning techniques for training LLMs.
\end{abstract}

\section{Introduction}

Large language models (LLMs)~\cite{radford2019language,brown2020language,openai2023gpt4,geminiteam2023gemini,touvron2023llama} have emerged as powerful tools for building human-level intelligence. It is crucial to understand the underlying principles and mechanisms in a more scientific way, for future development of the technologies and beyond.
One question we want to address is whether it is possible to formalize the typical training and inference processes of LLMs in a single and unified framework,
including pre-training, self-supervised fine-tuning (SFT)~\cite{ziegler2019fine,dodge2020fine}, reinforcement learning from human feedback (RLHF)~\cite{christiano2017deep,ouyang2022training,bai2022training}, in-context learning~\cite{min2022rethinking,lampinen2022can}, and chain-of-thought reasoning~\cite{zhou2022least,kojima2022large,wei2022chain,wang2022selfcot}. For example, pre-training is inherently language modeling, whereas RLHF is grounded in reinforcement learning, belonging to separate paradigms within machine learning.

In this position paper, we propose an agent perspective to unify the methodologies for training and improving LLMs. Our inspiration partly stems from a straightforward observation: LLMs like ChatGPT typically behave like an agent when responding to human users' requests, such as answering questions, providing advice, solving math problems, engaging in creative writing, or assisting with task completion.
The interactions between users and LLMs bear strong resemblance to a two-player game where player one (corresponding to the human users) and player two (corresponding to the LLMs) alternate in taking actions, with each attempting to maximize their individual internal goal. Indeed, this is a well-established subject that has been extensively studied in game theory (GT), reinforcement learning (RL), and multi-agent systems (MAS) with regards to creating agents that learn to play games.
One can refer to, e.g., the studies of learning-in-game in traditional GT literature~\cite{fudenberg1998theory,cressman2003evolutionary}, 
as well as the contemporary work that employs deep RL as the learning tools in multi-player games~\cite{heinrich2017reinforcement,brown2020combining,schmid2021search}, 
which is essentially multi-agent reinforcement learning (MARL)~\cite{shoham2008multiagent,lanctot2017unified}.

\begin{figure}
\centering
\includegraphics[width=0.9\linewidth]{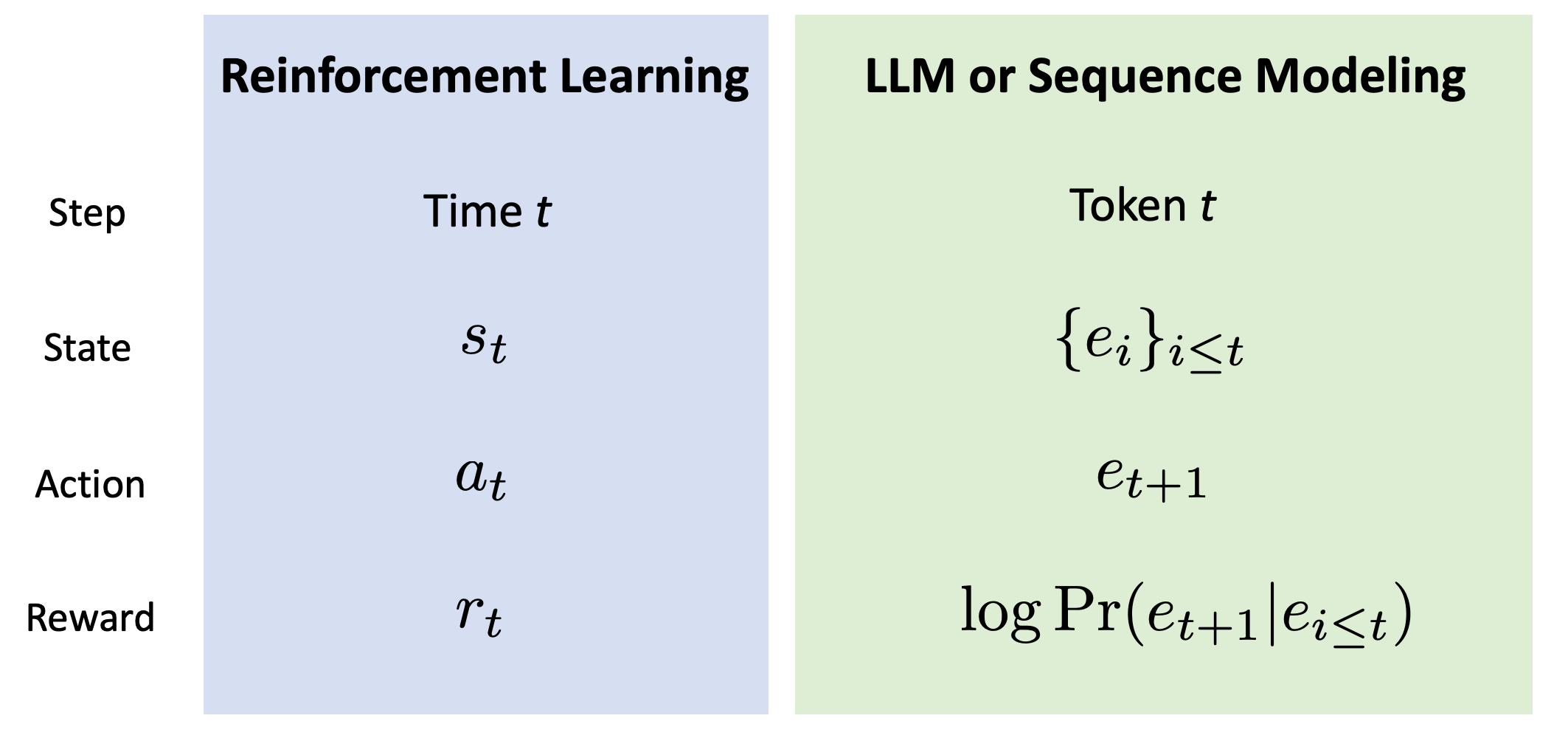}
\caption{\label{fig:connection} LLMs can be viewed as agents participating in language-based games in the framework of reinforcement learning.}
\vspace{-0.2in}
\end{figure}

In our setting, there are two agents or players who interact with each other by playing a language-based game over multiple turns. In each turn of the interaction, one player generates a sequence of tokens in multiple steps, based on the interactions between the two players up to that point. More concretely, we consider an RL formulation. Suppose that \emph{time} $t$ represents the step of generation made by player-one (user) or player-two (LLM); \emph{action} $a_t$  corresponds to generating the next token $e_{t+1}$; \emph{state} $s_t$ corresponds to the state of the interactions so far, which can simply be the sequence of tokens generated up to step $t$: $s_t = \{e_i\}_{i \leq t}$; the \emph{transition function} is defined to represent the transition to state $s_{t+1}$ given state $s_t$ and action $a_t$; \emph{reward} $r_t$ corresponds to the loss incurred at generation of token $e_{t+1}$. Player-one and player-two each have their own policies for taking actions or generating token sequences to maximize their expected cumulative rewards or expected returns. Player one and player two can each be formalized as an RL agent.
We summarize this formulation in~\Figref{fig:connection}.

In our framework, we conceptualize pre-training from a vast text corpus as behavior cloning of a sub-optimal policy of an ``average'' player-two from a large amount of log data of two-player language-based games (after proper processing). We regard SFT and RLHF as the respective methods for behavior cloning and policy learning aimed at developing an optimal policy for player-two. Within our framework, we offer explanations for various phenomena, including the learning of multiple tasks, chain-of-thought reasoning, prompting, hallucination, and in-context learning. We highlight that this new framework enables a comprehensive explanation and analysis of critical issues concerning the alignment of LLMs, such as vulnerabilities to attacks, leveraging the capabilities of LLMs, and the pursuit of superhuman intelligence. Additionally, our approach illuminates potential strategies for augmenting the capabilities of LLMs (player-two) through the refinement of data preparation and the advancement of learning methodologies.

Section~\ref{sec:formulation} defines the terms and notations necessary for our discussions. Section~\ref{sec:application} elaborates on how each training process of an LLM can be interpreted through the lens of a two-player game. Section~\ref{sec:insights} explores the implications arising from our framework established in Section~\ref{sec:application}. The same section also delves into new opportunities, challenges, and open questions. Finally, Section~\ref{sec:conclude} concludes our paper.

\section{Related Work}

Most relevant to us is the growing family of GPT models that have built the foundations of LLMs~\cite{radford2019language,brown2020language,openai2023gpt4,geminiteam2023gemini,touvron2023llama}. 
Thanks to advancements in technologies such as Transformer~\cite{vaswani2017attention} and language model training, the abundance of data, and the large scale of models, the current LLMs, with GPT-4 being a prominent example, have truly showcased human-level intelligence in numerous tasks
~\cite{bubeck2023sparks}.
There has been tremendous interest in understanding the training mechanisms.
Our position paper aims to spur further discussions on the research on LLMs.

SFT and RLHF~\cite{christiano2017deep,ziegler2019fine,dodge2020fine,liu2022few,ouyang2022training,bai2022training} are two prevalent training methods that steer LLMs to better align with human instructions or to generate outputs more in tune with human preferences. It is straightforward to consider the fine-tuning stage as behavior cloning and RLHF as reinforcement learning training. However, there has been no discussion about viewing pre-training as behavior cloning of a sub-optimal policy, to the best of our knowledge. Our discussions in this paper may provide profound insights into alignment techniques. For example, recent studies have investigated the possibility of self-aligning LLMs with guidance from another AI agent (see~\cite{lee2023rlaif,guo2024humaninstructionfree}). We are poised to offer novel guidelines for advancing these technologies.

Relevant to our scope are also methodologies that enhance LLMs through improved prompting, which include a broader discussion of prompt engineering~\cite{arora2022ask,chung2022scaling,white2023prompt}, chain-of-thought (CoT)~\cite{wei2022chain,wang2022selfcot,zhou2023leasttomost,kojima2023large}, and in-context learning~\cite{xie2021explanation,min2021metaicl,min2022rethinking,lampinen2022can,dong2022survey}, among others. As we will demonstrate, our formulation offers a novel perspective on these techniques.

Our study is closely related to the GT,  RL, and MAS literature on building agents to play multi-step games. 
In this body of work, the goal is to design learning algorithms that enable agents to achieve optimal policies, or, equivalently, to implement iterative equilibrium-finding procedures. The type of game can be either perfect information game~\cite{tesauro1995td,baxter2000learning,silver2017alphazero},
or imperfect information game, which may arise from simultaneous actions~\cite{littman1994markov,hu2003nash} or partial observability~\cite{vinyals2019grandmaster,brown2020combining}. 
In this study, we consider the interactions between humans and LLMs as games with action patterns similar to strategy card games~\cite{kowalski2023summarizing}, where efficient learning algorithms exist~\cite{xu2016convergence,grill2020monte,xi2023mastering}.

It has been discussed in the RL literature on its connection to the sequence modeling~\cite{wen2022multi}. For instance, a decision Transformer model is built for RL tasks~\cite{chen2021decision}. Offline RL algorithms using sequence modeling techniques are proposed by~\cite{janner2021offline}. Our paper provides a complementary view to this connection by establishing how the relevant RL literature on two-player games can offer new insights to the training of LLMs.

\section{Formulation}
\label{sec:formulation}

We now go through the preliminaries for a new framework explaining LLMs based on game theory, RL, and MAS. 

\begin{figure}
\centering
\begin{tikzpicture}
    \node at (0,0) {\includegraphics[width=0.45\linewidth]{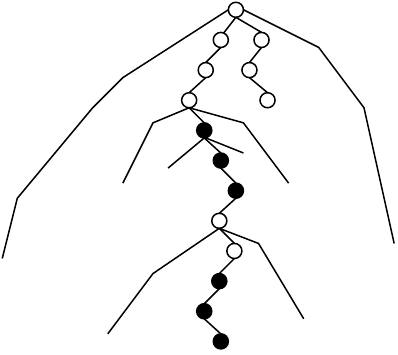}};
    \draw (-0.30,0.55) node[above] {\footnotesize{$s$}};
    \draw (-0.14,0.33) node[above] {\footnotesize{$a$}};
    \draw (0.3,0.23) node[above] {\footnotesize{$s'$}};
    \draw (-0.15,-1.49) node[above] {\footnotesize{$\tilde{s}$}};
    \draw (0.02,-1.85) node[above] {\footnotesize{$z$}};
\end{tikzpicture}
\caption{
The Game Tree representation for the LLM formulation.
Hollow circles: player one states, solid circles: player two states.
On the trajectory (episode) ending at terminal state $z$,
the representative states $s$, $s'$, $\tilde{s}$ and action $a$ are depicted,
where $s \sqsubset \tilde{s}$, $(s,a) \sqsubset \tilde{s}$, etc.
The active players are denoted as $P(s)=1$, $P(s')=2$, $P(\tilde{s})=2$.
The edge $(s,a)$ leads to $s'$ according to the transition function $s'=T(s,a)$.
The visiting probability $d^{\pi}(\tilde{s}|s,a)$,
starting from the edge $(s,a)$ and reaching the node $\tilde{s}$,
is given by \eqref{eq:cond-visit-prob}.
Similarly, $d^{\pi}(z)$ is given by \eqref{eq:decompose-s}.
}
\label{fig:gametree}
\vspace{-0.1in}
\end{figure}
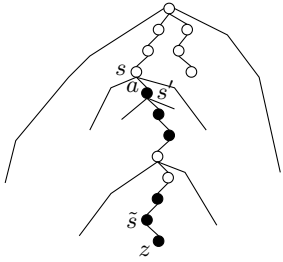

\mysubsubsection{Extensive Form Game and Game Tree}
We model the language interactions between a human and an LLM (agent) as an Extensive Form Game, where the two players interact in multi-turn and multi-step. 
For ease of discussion, we focus on the perfect information game setting. Later we will discuss the extension to the partially observable setting. 
Formally, let the game be $G=\left < \gN, \gS, \gZ, \gA, P, T, r \right >$, 
which can be represented with a \emph{Game Tree} as illustrated in~\Figref{fig:gametree}. 
Denote by $\gS$ the set of all possible states and state $s \in \gS$ corresponds to a node. 
The path from the root node to a node forms a trajectory.
A node and the trajectory to it are used interchangeably when the context is evident. 
Denote by $\gZ$ the set of \emph{terminal states} and terminal state $z \in \gZ$ corresponds to a leaf node. 
The action $a \in \gA(s)$, representing the generation of a token, is taken at $s$, 
where $\gA(s)$ denotes the vocabulary that is usually in the size of tens of thousands. 
Thus, $(s,a)$ corresponds to an edge. 
The \emph{transition function} $T$ can be either probabilistic or deterministic.
To model sequential token generation we adopt a deterministic function: $s'=T(s,a)=[s,a]$.
Applying it recursively, 
we can obtain a sequence of tokens and we denote a state $s$ using the sequence generated so far. 
Let $\gN=\{1, 2\}$ denote the set of indices for the two players. 
A player $i \in \gN$ can be either a human or an LLM.
At each $s$, the \emph{player function} $P(s)=i$ decides the \emph{active player}, 
i.e., it is player $i$'s turn to generate tokens. 
There exists a special token $\texttt{<eos>}$ to indicate the ending of generation for player $i$ in this turn. Subsequently, it becomes the other player's turn for token generation, and the process continues.
The multiple turns form an episode of a game, which is also a session of conversation.
To indicate that a node $s$ or an edge $(s,a)$ belongs to the trajectory $\tilde{s}$, 
we use the notation $s \sqsubset \tilde{s}$ or $(s,a) \sqsubset \tilde{s}$.
Finally, 
denote by ${\rm ch}(s)$ the set of trajectories originating from node $s$, 
and denote by ${\rm ch}(s,a)$ the set of trajectories originating from edge $(s,a)$.

\mysubsubsection{Policy and Visiting Probability}
At state $s$, 
the active player takes an action (generating a token) based on a conditional probability known as the \emph{policy}, 
denoted as $\pi^i(\cdot|s) \in \Delta(\gA(s))$, 
where the superscript $i$ denotes the active player $i=P(s)$
and $\Delta(\cdot)$ represents the probability distribution defined on the action set. 
The policy $\pi^i(\cdot|s)$ can be either \emph{deterministic} or \emph{probabilistic}, 
also referred to as \emph{pure} policy and \emph{mixed} policy in game theory, respectively.
Let $i$ represent one player, and $-i$ represent the other player. 
We refer to $\pi = (\pi^i, \pi^{-i})$ as a \emph{policy profile}. 
The policy $\pi^i(\cdot|\cdot)$ for player $i$ can be either predefined or learned. 
Once the policy profile is determined, 
the traversal of a game tree is given.
For a trajectory $s \in \gS$ or an episode $z \in \gZ$,
the \emph{Visiting Probability} at node $s$ or leaf $z$, 
denoted by $d^{\pi}(s)$ or $d^{\pi}(z)$, 
is simply defined to be the product of the players' policies along the trajectory:
\begin{equation}
    d^{\pi}(s)=\prod_{\forall (s',a') \sqsubset s} \pi^j(a'|s'), \quad d^{\pi}(z) = \prod_{\forall (s',a') \sqsubset z} \pi^j(a'|s'),
\label{eq:decompose-s}
\end{equation}
where the sum is over all the visited edges $(s',a')$ and the active player is determined on-the-fly by those visited states $j = P(s')$.
Reloading the notation $d^{\pi}(\cdot)$, 
we can also define the visiting probability for an edge $(s, a)$ as 
$d^{\pi}(s,a) \equiv d^{\pi}(s)\pi^j(a|s)$, where $j=P(s)$. 
Furthermore, 
we define the conditional visiting probability which gives the probability of traversing $\tilde{s}$ starting from node $s \sqsubset \tilde{s}$ or edge $(s, a) \sqsubset \tilde{s}$:
\begin{equation}
    d^{\pi}(\tilde{s}|s) \equiv d^{\pi}(\tilde{s})/d^{\pi}(s), \quad d^{\pi}(\tilde{s}|s,a) \equiv d^{\pi}(\tilde{s})/d^{\pi}(s,a).
\label{eq:cond-visit-prob}
\end{equation}
Note that in general the visiting probability $d^{\pi}(\cdot)$ depends on the policy profile $\pi = (\pi^i, \pi^{-i})$ from both players. 

\mysubsubsection{Reward, Return and Value Function}
\label{sec:reward-value}
Let player $i$ receive an \emph{immediate reward} $r^i(s,a) \in \mathbb{R}$ at edge $(s,a)$. 
Note that each player $i \in \gN$ has their own reward function $r^i(s,a)$, 
no matter whether state $s$ is $i$'s turn or not. 
For a trajectory $z \in \gZ$ and a player $i \in \gN$, 
we define the \emph{return} as his sum-of-rewards: 
$ R^i(z)=\sum_{\forall (s,a) \sqsubset z} r^i(s,a).$
In a slight abuse of notations, we also respectively define the \emph{expected return} and the \emph{state-action value function} as:
\begin{equation}
\label{eq:exp-return}
    R^i(\pi) = \mathbb{E}_{z \sim d^{\pi}(\cdot)} \left[ R^i(z) \right] = \sum_{z \in \gZ} d^{\pi}(z) R^i(z).
\end{equation}
\begin{align}
    Q^i_{\pi}(s,a) &= \mathbb{E}_{z \sim d^{\pi}(\cdot|s,a)} \left[ R^i(z) \right] \nonumber \\
                   &= \sum_{z \in \gZ, z \in {\rm ch}(s,a)}  d^{\pi}(z|s,a) R^i(z).
\label{eq:q-value}
\end{align}
It is worth mentioning that the expected return or value function of player $i$ also depends on the policy of the other player, 
i.e., the policy profile $\pi=(\pi^i, \pi^{-i})$.

The relationship between the returns $R^i(z)$ of the two players determines the type of the game. In a two-player game,
$R^1(z) + R^2(z) = 0$ represents a zero-sum game, which is purely adversarial;
$R^1(z) - R^2(z) = 0$ indicates an identical interest game, which is purely cooperative;
$R^1(z) + R^2(z) = c(z)$ represents a general case, which is mixed competitive-cooperative (collaborating to make the cake bigger but conflicting when dividing the cake).

\begin{figure}
\centering

\begin{tikzpicture}
    \node at (0,0) {\includegraphics[width=1.0\linewidth]{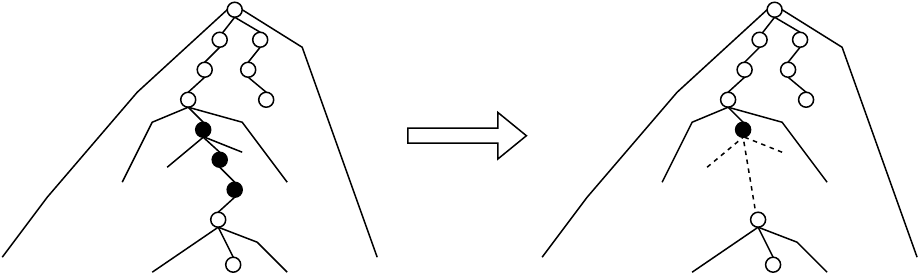}};

    \draw (-2.60,0.18) node[above] {$s$};
    \draw (-2.5,-0.05) node[above] {$a$};
    \draw (-2.4,-0.95) node[above] {$s'$};

    \draw (2.22,0.18) node[above] {$s$};
    \draw (2.37,-0.05) node[above] {$a$};
    \draw (2.43,-0.95) node[above] {$s'$};
\end{tikzpicture}

\caption{
A fixed player $-i$ policy induces an MDP to player $i$,
where the environmental dynamic is ${\rm Pr(s'|s,a)=d^{\pi}(s'|s,a)}$ 
by using the visiting probability \eqref{eq:cond-visit-prob}
and by noting that the underlying transition $T(\cdot,\cdot)$ is deterministic.
}
\label{fig:reduce_rl}
\vspace{-0.1in}
\end{figure}
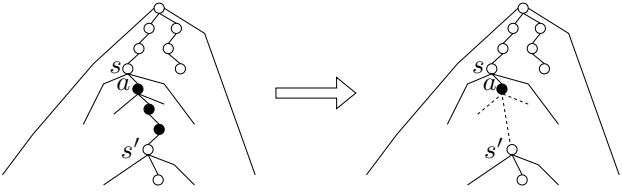

\mysubsubsection{Reinforcement Learning}
Suppose that we aim to learn the policy $\pi^i$ for player $i$ while the other player's policy $\pi^{-i}$ is fixed. In this scenario, a \emph{Markov Decision Process (MDP)} is induced from the perspective of player $i$, where the \emph{environment} is determined by the other player's policy $\pi^{-i}$ and the transition function $T(\cdot,\cdot)$, 
as illustrated by \Figref{fig:reduce_rl}.
The objective for player $i$ is to maximize
\begin{equation}
\label{eq:max-exp-return}
    \max_{\pi^i} R^i(\pi) \equiv \max_{\pi^i} R^i(\pi^i, \pi^{-i}),
\end{equation}
which corresponds to a (single-agent) \emph{Reinforcement Learning} problem for player $i$.

\mysubsubsection{Solution Concept and Nash Equilibrium}
When both players' policies are subject to learning, the optimal policy profile $\pi^{*}=(\pi^{i,*}, \pi^{-i,*})$ is referred to as a \emph{Solution Concept}. One widely used solution concept in game theory is the \emph{Nash Equilibrium} (NE), which represents a \emph{fixed point} in the policy space that satisfies the following condition:
\begin{equation}
\label{eq:ne1}
R^i(\pi^{i,*}, \pi^{-i,*}) \geq R^i(\pi^{i}, \pi^{-i,*}),\quad \forall \pi^{i}, \enspace \forall i \in \gN.
\end{equation}
This condition indicates that if both players adopt NE policies, neither player can achieve a higher expected return by unilaterally altering his policy.

\section{Interpretations}
\label{sec:application}

In this section, 
we interpret various LLM technologies using the theoretic framework in the previous section.

\subsection{Overview}
Suppose that in the discussions in section \ref{sec:pretrain}-\ref{sec:rlhf}, 
the player-one (\emph{human})'s policy is fixed such that 
it reduces to a single-agent RL problem for player-two (\emph{LLM}). 
We sometimes rewrite the notations by omitting the player index superscript $i=2$ for simplicity. Denote by $\pi_{\theta}(a_t|s_t) \equiv \pi^{i}(a_t|s_t)$ the policy over state $P(s_t)=i$ with parameter $\theta$,
denote by $J(\theta) = R^i(\pi) \equiv R^i(\pi_{\theta},\pi^{-i})$ the objective function in~\eqref{eq:exp-return}, and denote by $Q_{\pi_{\theta}}(s_t,a_t) \equiv Q_{\pi}^i(s_t,a_t) \equiv Q_{(\pi^{i},\pi^{-i})}^i(s_t,a_t)$ the state-action value in~\eqref{eq:q-value}. 

\subsection{Pre-training}
\label{sec:pretrain}

\begin{figure}
\includegraphics[width=1.0\linewidth]{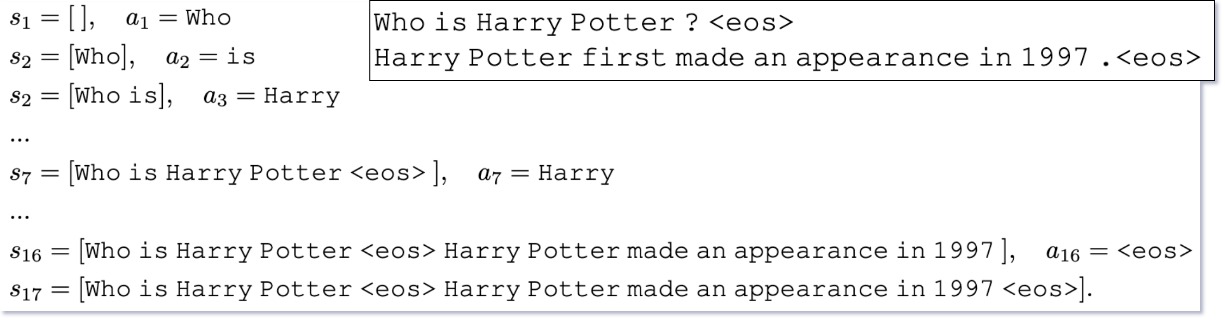}
\caption{Example of token sequence in pre-training data. It can be interpreted as a series of state-action pairs resulting from the game of two RL agents.}
\label{fig:eg1}
\vspace{-0.1in}
\end{figure}

We regard pre-training of LLM as \emph{behavior cloning} of the average player-two's policy, which is sub-optimal when aiming to construct an ideal player-two. Nonetheless, the data in the form of documents, approximating an extensive collection of logs from two-player games, can be highly valuable for our purpose, primarily due to its vast scale. After pre-training, the LLM becomes a Transformer-based policy model capable of taking actions or generating token sequences for player-two in two-player games. 

The data in pre-training is not explicitly formatted to depict the sequential actions of two players. We posit that the inclusion of specific delimiters, akin to tags in HTML documents, could serve as cues to segment a token sequence into two parts. The first part would represent the actions of the first player, who poses a question, while the second part would represent the actions of the second player, who provides a response. The utility of delimiters within pre-training data is pointed out in previous work~\cite{brown2020language}.

Consider a document comprised of a token sequence as depicted in~\Figref{fig:eg1}, which is divided into two segments: a question and an answer. The corresponding sequence of states and actions for the document can be delineated as shown in~\Figref{fig:eg1}. 
Note that there are a total of $T=16$ actions and $T + 1 =17$ states. 

Specifically, the LLM is trained on a very large corpus of documents $\mathcal{D}$, to predict the next token in an auto-regressive manner by optimizing the log-likelihood (equivalent to minimizing the negative log-likelihood):
\begin{equation}\label{eq:pretrain-loss}
\hat{\pi} = \argmax_{\pi}\tilde{\mathbb{E}}_{(s_t,a_t) \sim \mathcal{D}} \left[ \log \pi(a_t | s_t) \right]
\end{equation}
where $\tilde{\mathbb{E}}$ denotes batch-level averaging. 
When calculating the pre-training loss for the data depicted in Figure \ref{fig:eg1}, it is necessary to include all pairs of states and actions $(s_t,a_t)$ for each $t$ ranging from 1 to 16.

\subsection{SFT}
\label{sec:sft}
We perceive supervised fine-tuning (SFT) \cite{ziegler2019fine,dodge2020fine} of LLM as behavior cloning of the optimal policy for player-two. Training data is treated as logged interactions between the two players, with player-two being the agent we aim to construct. The questions (data) are interpreted as actions (token sequences) from player-one, while the answers are interpreted as actions (token sequences) from player-two. The goal is for the LLM to learn from the demonstrations of the ideal player-one, which are typically annotated by human experts. The model is trained by maximizing the log-likelihood of the data $\mathcal{D}$
\begin{equation}
    \pi^* = \argmax_{\pi}\tilde{\mathbb{E}}_{(s_t,a_t) \sim \mathcal{D}} \left[ \log \pi(a_t | s_t) \right],
    \label{eq:sft-loss}
\end{equation}
where $\tilde{\mathbb{E}}$ stands for batch-level averaging.

\begin{figure}
\includegraphics[width=1.0\linewidth]{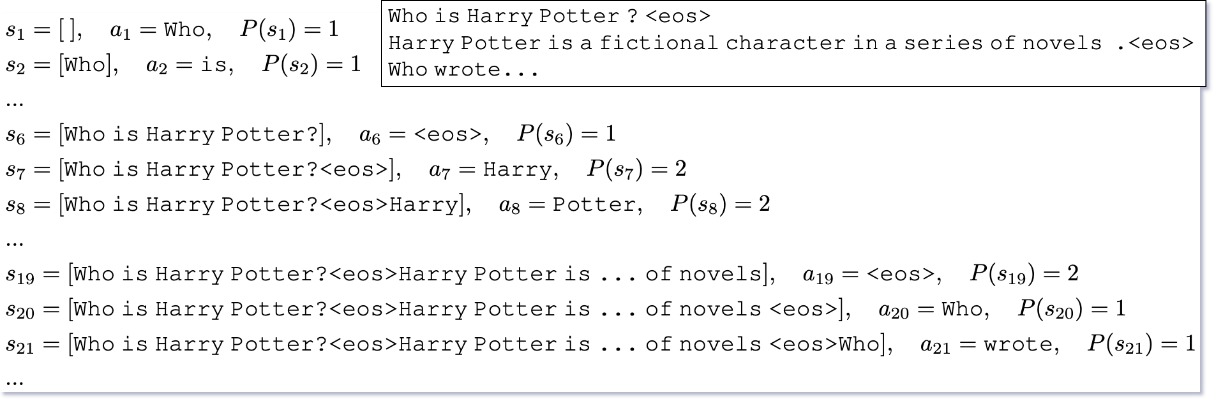}
\caption{Example of token sequence in SFT data. It can be interpreted as a series of state-action pairs resulting from the game of two RL agents.}
\label{fig:eg2}
\end{figure}

Suppose that we are provided with an example of SFT data, as in~\Figref{fig:eg2}. The states, actions and player functions can be then specified as in the figure. This figure includes sequences of tokens from both player-one and player-two. 
During player-two's turn, 
state $s_7$ serves as the starting point, 
and state $s_{19}$ marks the end. 
Player-two engages in a series of actions, which leads to a progression of state and action pairs $(s_t, a_t)$ for $t=7,8,...,18,19$. 
The token generation process adheres to equation~\ref{eq:sft-loss}. 
The action sequence $[a_1,...,a_6]$ corresponds to the question,
and the subsequent action sequence $[a_7,...,a_{19}]$ corresponds to the answer in the SFT data.

\subsection{RLHF}
\label{sec:rlhf}

In RLHF, the LLM undergoes further fine-tuning in a two-stage process~\cite{christiano2017deep,bai2022constitutional,ouyang2022training,lee2023rlaif}. A reward model, $r^{i}(s,a)$ or $R^{i}(z)$, is first trained based on data from human preferences. The LLM after SFT is then further trained using policy gradient based RL, e.g., the PPO algorithm~\cite{schulman2017proximal}. 
RLHF is considered as a single agent RL process aimed at enhancing the LLM's capability to take actions or generate tokens, effectively emulating an ideal player-two.

The objective of RL is to maximize the expected return as in \eqref{eq:max-exp-return}. 
This can be done by gradient ascent~\cite{sutton1999policy} that explicitly gives the gradient:
\begin{equation}
    \nabla_{\theta} J(\theta) \approx \mathbb{\tilde{E}}_{(s_t,a_t) \sim \mathcal{D}} 
        \left[  Q_{\pi_{\theta}}(s_t,a_t) \nabla_{\theta} \log \pi_{\theta}(a_t|s_t)  \right],
\label{eq:policy-gradient}
\end{equation}
where $\mathcal{D}$ is a ``temporary dataset'' consisting of the fresh rollout data for player two sampled by the current policy profile $(\pi_{\theta},\pi^{-i})$, i.e., the on-policy RL. 
The PPO algorithm is similar in essence to~\eqref{eq:policy-gradient}, but incorporates more advanced safe stepping techniques and variance reduction techniques.

\subsection{Additional Settings}
\label{sec:additional}
We explore several other setups of LLMs as two-player games.

\mysubsubsection{Meta Learning}In all three-stages of LLM training, the dataset $\mathcal{D}$ in fact contains a large number of subsets that feature different tasks and different transition dynamics (${\cal D}_i,i=1,2,\cdots,n$). 
Therefore, the formulations in sections 4.1-4.3 can also be viewed as  a \emph{meta policy learning} process \cite{nagabandi2018learning,gupta2018meta,rakelly2019efficient}, in which player-two learns a meta policy from the entire dataset of different tasks.

During the three stages of LLM training, the dataset $\mathcal{D}$ actually comprises a multitude of subsets, representing different tasks with their different RL configurations ($D_i, i=1,2,\cdots,n$):
\[
\mathcal{D} = \mathcal{D}_1 \cup \mathcal{D}_2 .... \cup \mathcal{D}_n
\]
Consequently, the approaches outlined in sections 4.1-4.3 can be interpreted as a meta policy learning process, 
in which player-two acquires a meta policy that is applicable across a large number of tasks represented by the entire dataset:
\begin{equation}
    \pi^* = \argmax_{\pi}~\tilde{\mathbb{E}}_{G \in \mathcal{D}} \left[\tilde{\mathbb{E}}_{(s_t,a_t) \sim \mathcal{D}(G)} \left[ \log \pi(a_t | s_t) \right]\right]
    \label{eq:pretrain-loss-meta}
\end{equation}

In the notation previously introduced, a game formulation $G_i$ is first selected from a distribution of games, and then a set of episodes is drawn from the episode distribution $\mathcal{D}_i$ of the game $G_i$. This analogy further clarifies the often-noted zero-shot or few-shot learning capabilities exhibited by LLMs such as GPT-3 and others. %

\mysubsubsection{Partially Observed States}
A natural extension of our framework is to conceptualize the games between two players such that the states of environment are only partially observed. From the perspective of either player-one or player-two, the scenario resembles a partially observed Markov decision process (POMDP). In a competitive setting, player-one might deliberately conceal information during its engagement with player-two. Even in a collaborative or mixed scenario, it is possible that player-one does not adequately communicate all the essential details~\cite{andreas2022language}. Under these circumstances, player-two must deduce the true states of the environment.

The POMDP extension helps us better comprehend the significance of the quality of prompts, i.e., the questions or extra elucidations provided by player-one. A more effectively articulated prompt diminishes the uncertainty and noise within the token sequence from player-one, enabling player-two to precisely discern the states and execute appropriate actions~\cite{zhou2022large,white2023prompt}.

The extension also enables us to understand the efficacy of chain-of-thought (CoT) reasoning~\cite{wei2022chain,wang2022selfcot}. In this approach, player-two produces a token sequence that reflects its ``thought process'', that is, the CoT, prior to generating the token sequence that serves as a response to player-one. Player-two can utilize the observation to deduce the states while deciding on the subsequent actions. The information provided by the CoT can offer a clearer indication of the true states and lead to better actions.

The POMDP perspective additionally compels us to contemplate a more comprehensive formulation for the training of LLMs, which we regard as future work. This opens up new prospects for constructing world models for both player-two and player-one, thereby enhancing their capacity for reasoning and planning~\cite{ha2018world,hao2023reasoning}. However, to fully leverage the players' ability to create and use their world models, it would necessitate capabilities for sensing and gathering data across multiple modalities, encompassing image, video, speech, and others.

\mysubsubsection{Hallucination}
The two-player games in question are ``language-based games'', 
where the players must engage using human language. This implies that the token sequences produced by both players are entirely legitimate in human language. Put differently, these token sequences are required to adhere to the syntactic, semantic, and pragmatic norms of the language, which concurrently constitute the rules of the games. An underlying premise of this two-player game formulation is that the rules are not explicitly provided; instead, they must be inferred from the data obtained in the past games by many different players.

The process of learning, especially pre-training of LLMs, can largely fulfill the goal of discerning the rules of the game from vast amount of text data. It seems that the token sequences generated by LLMs are syntactically and pragmatically impeccable. Nevertheless, the same cannot be said for their semantic accuracy. At times, LLMs may produce token sequences that seem credible yet lack factual substantiation - a phenomenon known as \emph{hallucination} \cite{zhang2023siren,li2023halueval,ji2023survey,liu2023trustworthy}. It is evident that the learning mechanisms of LLMs are not fully equipped to address the issue of hallucination effectively.

We posit that the primary cause of hallucination in LLMs is their lack of construction and connection to world models~\cite{lecun2022path}. Although LLMs can generate responses to fact-based questions, showing that they store knowledge in the form of language, they lack world models that could help verify whether their statements are factually correct. Consequently, this leads to the breach of semantic rules pertaining to the veracity of the statements made.

\mysubsubsection{In-Context Learning}As outlined in the GPT-3 paper~\cite{brown2020language}, a substantial volume of question-answer pairs can be assembled, annotated, and incorporated into the pre-training of an LLM through what is called in-context learning \cite{min2021metaicl,dong2022survey}. This approach can be regarded as an imitation of average player-two within our framework. Meanwhile, the learning is similar to off-policy reinforcement learning and is conducted in a mini-batch fashion. With this kind of training, an LLM is capable of executing zero-shot or few-shot learning effectively, without the need of additional parameter tuning. During inference, the LLM can be immediately utilized in an in-context learning scenario to enhance the capability of player-two \cite{zhu2023transfer}, where a handful of question-answer examples are given as a prompt.

\subsection{Alignment}

To improve the capabilities of LLMs (player-two), it is advantageous to also consider the capabilities of the environment (player-one) and the overall game configuration. This will foster the creation of novel alignment algorithms that go beyond the reinforcement learning from human feedback (RLHF). We can explore various two-player game settings as outlined below, which lead to solutions for a range of complex alignment challenges. These include adversarial attacks on LLMs, harnessing the full potential of LLMs, and advancing LLMs towards superhuman intelligence, corresponding with adversarial player, cooperative player, and the Nash equilibrium, respectively.

The alignment research is poised for significant advancement through the development of new algorithms within our framework, moving beyond the current PPO algorithm and the recently introduced direct preference optimization (DPO) algorithm~\cite{rafailov2023direct}. We also believe that the large collection of learning algorithms developed for games, including those tailored for specific games~\cite{guo2021adversarial},
as well as those created for imperfect games~\cite{kahn2017uncertainty,wang2020reinforcement,wang2021policy,perolat2021poincare,an2021uncertainty,sokota2022unified,zhang2023uncertainty}, 
can also inspire the design of new player-two algorithms that could significantly enhance the alignment of LLMs~\cite{wang2022self,sun2023principle,guo2024humaninstructionfree}.

\subsection{Adversarial Players}
Enhancing the trustworthiness of LLMs is crucially important in the development of LLMs, including alignment. From the viewpoint of a two-player game, all the issues arise due to an adversarial player-one. These include but are not limited to reliability, safety, robustness, fairness, adherence to social norms, and resistance to misuse~\cite{liu2023trustworthy,wang2023decodingtrust}. The key question here is how to model player-one. One possible approach is to train another LLM as player-one, and let player-one and player-two play against each other. 
In learning, they can enhance their capabilities through a large number of game-plays, analogous to the self-play RL in AlphaZero~\cite{silver2017alphazero}.

Recent research on ``red teaming''~\cite{perez2022red,ganguli2022red} is based on this presumption. Within our game-theoretic framework, player-one and player-two are participants in a zero-sum game $R^1(z)+R^2(z)=0$, 
indicating a purely competitive interaction (cf., \secref{sec:reward-value}). 
With adequate training, player-one becomes adept at creating a range of challenging questions designed to stump player-two, while player-two simultaneously learns to respond to these questions safely and effectively. Their skills, or their policy profile, improve over time, ultimately converging to a Nash equilibrium as in~\eqref{eq:ne1}.

\subsection{Cooperative Players}

The majority of interactions between an LLM and human users can be seen as cooperative games involving player-two and player-one \cite{nash1953two,claus1998dynamics}. This encompasses interactions like responding to queries, offering guidance, providing emotional health assistance, entertaining, and helping with various tasks by the LLM.

A critical technique in employing LLMs involves the art of prompting. It is broadly recognized that effective prompting can remarkably improve the quality of responses by LLMs (player-two). This can be viewed as augmenting player-one's questions with additional explanations that act as context, enabling player-two to respond more appropriately. 
The reason can be interpreted as that without proper prompting, the intrinsic states of player-one are not fully exposed in its generated tokens.

An important question is how to enhance prompting (additional actions of player-one) via a learning mechanism. In~\cite{saunders2022self,bai2022constitutional}, 
a predetermined policy is applied to player-one, whereas a policy for player-two is formulated through learning. 
It is also feasible to consider a two-player identical interest game $R^1(z)=R^2(z)$, as discussed in~\secref{sec:reward-value}.
In such a cooperative scenario, the player-one policy should be carefully initialized and adjusted, serving as a dedicated prompt provider, 
and both the player-one and player-two policies are learnable, 
being refined over time and ultimately approaching NE as in~\eqref{eq:ne1}.

\subsection{Superhuman Intelligence}

A recently defined objective within the community, including organizations like OpenAI, is to align LLMs for tasks that require superhuman intelligence. Currently, there are many open questions on the feasibility of accomplishing this formidable goal. OpenAI's newly published work~\cite{burns2023weak} explores the concept of employing less powerful models to guide a much more stronger LLMs in executing tasks of superhuman complexity. Further investigation is essential for this challenging issue.

If we consider alignment involving two agents capable of self-play and simultaneous improvement, it appears to be a more natural approach to attaining superhuman intelligence. This concept is not novel; we have already seen AlphaZero's triumph over top human Go players through self-play between two AI agents.

If we consider both players as learning agents progressing towards an equilibrium strategy, we must ponder what this equilibrium would mean for an LLM. Would the equilibrium strategy lead to a superhuman level of policy? Exploring these questions will yield both theoretical and empirical insights for the creation of LLMs.

\section{Insights and Open Questions}
\label{sec:insights}

In this section, we offer several insights derived from our framework and pose several questions for future research.

\subsection{Data Preparation}
Our framework offers valuable insights for data preparation. For example, the game approach to the pre-training phase suggests that data pre-processing methods which convert unstructured text data into a ``question-answer" (Q-A) structure reflecting the actions of two players can facilitate the training of the learning player-two (LLM), as indicated by \cite{brown2020language}.

Furthermore, examination of chain-of-thought (CoT) suggests that the pre-training data may already include a ``question-reasoning-answer" (Q-C-A) structure from which the model can learn. It is possible that pre-trained LLMs have acquired the CoT capability, which can be effectively brought out through proper prompt engineering.

These two observations lead to the critical question of whether explicitly structuring data into Q-A or Q-C-A formats could enhance the training process even further. We leave this question to the research community.

Moreover, it is apparent that the success of SFT and RLHF is contingent on the policy of player-one. The prevalent method of data gathering for SFT and RLHF resembles the process of obtaining samples from the policy of an "average" player-one, which is essentially data sourced from typical users. We anticipate that our framework will motivate a re-evaluation of the dataset preparation for SFT and RLHF. This could involve a more deliberate modeling and training of player-one, aimed at producing higher quality data \cite{zhu2023unmasking}. This data would then be used by player two for learning of an optimal policy. The concept shares similarities with red teaming; however, we consider here a more principled approach for the collection of superior data.

\subsection{Training Methods}

\mysubsubsection{Reward Functions}
The two-player game approach also provides fresh insights into reinforcement learning of LLMs, including RLHF. By substituting the reward function with one that represents the two-player games, we can expand the versatility of reinforcement learning. For example, by devising a reward function that embodies a zero-sum game, it is conceivable to train a model to learn to refuse to respond to particular queries. This area of study, recently gaining interest under the theme of ``LLM unlearning'' \cite{yao2023large,eldan2023s}, highlights the potential of reward function design in reinforcement learning.

\mysubsubsection{Value Functions} Incorporation of value functions into the training of LLMs and the users they interact with can also be explored.
They could enhance the learning of policies for both LLMs and the users, corresponding to player-two and player-one respectively.
Given our framework, investigating this topic is of significant importance.

Our framework also underscores the significance of formalizing and employing value functions for long-term planning. Although the learning of LLMs have inherently included the learning of language-based games within their training regime, they lack the notion of a long-term value function, which is a standard component in the training of reinforcement learning agents for long-term planning \cite{guo2014deep,gupta2019relay, schrittwieser2020mastering}. The explicit computation of long-term value functions could aid the LLMs exploit "long chains of reasoning," thereby reducing the effect of hallucination and enhancing the models' reasoning and planning abilities.

\mysubsubsection{Multi-Agents} The recent surge in interest centers on collaborations among multiple LLMs, with the objective of enhancing the overall performance of the models. Research on LLM debates, for instance, has offered supporting evidence for this collaborative approach \cite{du2023improving,chan2023chateval}. Such results highlight the prospective benefits of utilizing multi-agent technologies during the training of LLMs to boost their collective capabilities.

\mysubsubsection{Learning from Scratch} Recent research demonstrates that with a simple win-or-lose reward for terminal states, an agent can master the board games like Go, Chess, and Shogi. This success is attained a unified approach to multi-agent reinforcement learning (MARL) that learns from the scratch~\cite{silver2017mastering,silver2017alphazero}.
Consequently, one might naturally ask: given an omniscient reward function/model,  can the players learn, \emph{tabula rasa}, human-level language-based ``game'' abilities?  It is noteworthy that in our formulation of a perfect information game for two-players, a communication channel with unlimited capacity among the players is readily feasible. Furthermore, the upper bound of the gaming capability is subject to the reward function/model. Is it possible to establish a reward function or model that facilitates the acquisition of human-level abilities?

\mysubsubsection{Learning in World Environment}
It is widely held that natural language has evolved from interactions among humans~\cite{pinker2007stuff}. Initial efforts to replicate this phenomenon have been made in controlled laboratory settings. It is stated in ~\cite{mordatch2018emergence}: ``\emph{...teach AI agents to create language by dropping them into a set of simple worlds, giving them the ability to communicate, and then giving them goals that can be best achieved by communicating with other agents}'' (quoted from~\cite{learning-to-communicate}).

The conceptualization of LLMs as players in a game, or more broadly, as agents, might allow us to expand the range of AI agents. This expansion could take us from agents that learn language alone to those that learn language within a multimodal world environment, and from agents that solely learn a language model to those that learn an integrated system of language and multimodal models~\cite{lecun2022path}.

\section{Concluding remarks}
\label{sec:conclude}
This position paper presents a two-player game framework to re-evaluate the training, tuning, and alignment processes of LLMs. We provide a detailed argument for viewing each training process of an LLM through the lens of a two-player game. Our formulation provides implications and insights that may explain the current successes of LLMs, including the celebrated SFT, RLHF, and in-context learning paradigms. Our formulation suggests potential future developments, including possible ways to more effectively align an LLM to follow human preferences, and potential ways to improve reasoning by more explicit modeling of long-term values. We hope that our paper will inspire more discussions between the LLM and game theory, RL and MAS communities.

\section{Impact Statements}

This paper presents a two-player game formulation to explain some of the underlying mechanisms of LLMs. We believe our paper has the potential to unveil the inner workings behind the observed successes and failures of LLMs. The comprehensive understanding and insights derived from our paper could enhance the trustworthiness of LLMs. For example, we provide recommendations for comprehending the hallucination aspect of LLMs. Additionally, our paper offers suggestions for enhancing the capabilities of LLMs. For instance, our discussion on developing reward functions that embody different zero-sum games has the potential to inspire new alignment algorithms and applications. %

\bibliographystyle{icml2024}
\bibliography{ref}

\begin{thebibliography}{95}
\providecommand{\natexlab}[1]{#1}
\providecommand{\url}[1]{\texttt{#1}}
\expandafter\ifx\csname urlstyle\endcsname\relax
  \providecommand{\doi}[1]{doi: #1}\else
  \providecommand{\doi}{doi: \begingroup \urlstyle{rm}\Url}\fi

\bibitem[An et~al.(2021)An, Moon, Kim, and Song]{an2021uncertainty}
An, G., Moon, S., Kim, J.-H., and Song, H.~O.
\newblock Uncertainty-based offline reinforcement learning with diversified
  q-ensemble.
\newblock \emph{Advances in neural information processing systems},
  34:\penalty0 7436--7447, 2021.

\bibitem[Andreas(2022)]{andreas2022language}
Andreas, J.
\newblock Language models as agent models.
\newblock \emph{arXiv preprint arXiv:2212.01681}, 2022.

\bibitem[Arora et~al.(2022)Arora, Narayan, Chen, Orr, Guha, Bhatia, Chami,
  Sala, and R{\'e}]{arora2022ask}
Arora, S., Narayan, A., Chen, M.~F., Orr, L., Guha, N., Bhatia, K., Chami, I.,
  Sala, F., and R{\'e}, C.
\newblock Ask me anything: A simple strategy for prompting language models.
\newblock \emph{arXiv preprint arXiv:2210.02441}, 2022.

\bibitem[Bai et~al.(2022{\natexlab{a}})Bai, Jones, Ndousse, Askell, Chen,
  DasSarma, Drain, Fort, Ganguli, Henighan, et~al.]{bai2022training}
Bai, Y., Jones, A., Ndousse, K., Askell, A., Chen, A., DasSarma, N., Drain, D.,
  Fort, S., Ganguli, D., Henighan, T., et~al.
\newblock Training a helpful and harmless assistant with reinforcement learning
  from human feedback.
\newblock \emph{arXiv preprint arXiv:2204.05862}, 2022{\natexlab{a}}.

\bibitem[Bai et~al.(2022{\natexlab{b}})Bai, Kadavath, Kundu, Askell, Kernion,
  Jones, Chen, Goldie, Mirhoseini, McKinnon, et~al.]{bai2022constitutional}
Bai, Y., Kadavath, S., Kundu, S., Askell, A., Kernion, J., Jones, A., Chen, A.,
  Goldie, A., Mirhoseini, A., McKinnon, C., et~al.
\newblock Constitutional ai: Harmlessness from ai feedback.
\newblock \emph{arXiv preprint arXiv:2212.08073}, 2022{\natexlab{b}}.

\bibitem[Baxter et~al.(2000)Baxter, Tridgell, and Weaver]{baxter2000learning}
Baxter, J., Tridgell, A., and Weaver, L.
\newblock Learning to play chess using temporal differences.
\newblock \emph{Machine learning}, 40:\penalty0 243--263, 2000.

\bibitem[Brown et~al.(2020{\natexlab{a}})Brown, Bakhtin, Lerer, and
  Gong]{brown2020combining}
Brown, N., Bakhtin, A., Lerer, A., and Gong, Q.
\newblock Combining deep reinforcement learning and search for
  imperfect-information games.
\newblock \emph{Advances in Neural Information Processing Systems},
  33:\penalty0 17057--17069, 2020{\natexlab{a}}.

\bibitem[Brown et~al.(2020{\natexlab{b}})Brown, Mann, Ryder, Subbiah, Kaplan,
  Dhariwal, Neelakantan, Shyam, Sastry, Askell, et~al.]{brown2020language}
Brown, T., Mann, B., Ryder, N., Subbiah, M., Kaplan, J.~D., Dhariwal, P.,
  Neelakantan, A., Shyam, P., Sastry, G., Askell, A., et~al.
\newblock Language models are few-shot learners.
\newblock \emph{Advances in neural information processing systems},
  33:\penalty0 1877--1901, 2020{\natexlab{b}}.

\bibitem[Bubeck et~al.(2023)Bubeck, Chandrasekaran, Eldan, Gehrke, Horvitz,
  Kamar, Lee, Lee, Li, Lundberg, Nori, Palangi, Ribeiro, and
  Zhang]{bubeck2023sparks}
Bubeck, S., Chandrasekaran, V., Eldan, R., Gehrke, J., Horvitz, E., Kamar, E.,
  Lee, P., Lee, Y.~T., Li, Y., Lundberg, S., Nori, H., Palangi, H., Ribeiro,
  M.~T., and Zhang, Y.
\newblock Sparks of artificial general intelligence: Early experiments with
  gpt-4, 2023.

\bibitem[Burns et~al.(2023)Burns, Izmailov, Kirchner, Baker, Gao,
  Aschenbrenner, Chen, Ecoffet, Joglekar, Leike, et~al.]{burns2023weak}
Burns, C., Izmailov, P., Kirchner, J.~H., Baker, B., Gao, L., Aschenbrenner,
  L., Chen, Y., Ecoffet, A., Joglekar, M., Leike, J., et~al.
\newblock Weak-to-strong generalization: Eliciting strong capabilities with
  weak supervision.
\newblock \emph{arXiv preprint arXiv:2312.09390}, 2023.

\bibitem[Chan et~al.(2023)Chan, Chen, Su, Yu, Xue, Zhang, Fu, and
  Liu]{chan2023chateval}
Chan, C.-M., Chen, W., Su, Y., Yu, J., Xue, W., Zhang, S., Fu, J., and Liu, Z.
\newblock Chateval: Towards better llm-based evaluators through multi-agent
  debate, 2023.

\bibitem[Chen et~al.(2021)Chen, Lu, Rajeswaran, Lee, Grover, Laskin, Abbeel,
  Srinivas, and Mordatch]{chen2021decision}
Chen, L., Lu, K., Rajeswaran, A., Lee, K., Grover, A., Laskin, M., Abbeel, P.,
  Srinivas, A., and Mordatch, I.
\newblock Decision transformer: Reinforcement learning via sequence modeling.
\newblock \emph{Advances in neural information processing systems},
  34:\penalty0 15084--15097, 2021.

\bibitem[Christiano et~al.(2017)Christiano, Leike, Brown, Martic, Legg, and
  Amodei]{christiano2017deep}
Christiano, P.~F., Leike, J., Brown, T., Martic, M., Legg, S., and Amodei, D.
\newblock Deep reinforcement learning from human preferences.
\newblock \emph{Advances in neural information processing systems}, 30, 2017.

\bibitem[Chung et~al.(2022)Chung, Hou, Longpre, Zoph, Tay, Fedus, Li, Wang,
  Dehghani, Brahma, Webson, Gu, Dai, Suzgun, Chen, Chowdhery, Castro-Ros,
  Pellat, Robinson, Valter, Narang, Mishra, Yu, Zhao, Huang, Dai, Yu, Petrov,
  Chi, Dean, Devlin, Roberts, Zhou, Le, and Wei]{chung2022scaling}
Chung, H.~W., Hou, L., Longpre, S., Zoph, B., Tay, Y., Fedus, W., Li, Y., Wang,
  X., Dehghani, M., Brahma, S., Webson, A., Gu, S.~S., Dai, Z., Suzgun, M.,
  Chen, X., Chowdhery, A., Castro-Ros, A., Pellat, M., Robinson, K., Valter,
  D., Narang, S., Mishra, G., Yu, A., Zhao, V., Huang, Y., Dai, A., Yu, H.,
  Petrov, S., Chi, E.~H., Dean, J., Devlin, J., Roberts, A., Zhou, D., Le,
  Q.~V., and Wei, J.
\newblock Scaling instruction-finetuned language models, 2022.

\bibitem[Claus \& Boutilier(1998)Claus and Boutilier]{claus1998dynamics}
Claus, C. and Boutilier, C.
\newblock The dynamics of reinforcement learning in cooperative multiagent
  systems.
\newblock \emph{AAAI/IAAI}, 1998\penalty0 (746-752):\penalty0 2, 1998.

\bibitem[Cressman(2003)]{cressman2003evolutionary}
Cressman, R.
\newblock \emph{Evolutionary dynamics and extensive form games}, volume~5.
\newblock MIT Press, 2003.

\bibitem[Dodge et~al.(2020)Dodge, Ilharco, Schwartz, Farhadi, Hajishirzi, and
  Smith]{dodge2020fine}
Dodge, J., Ilharco, G., Schwartz, R., Farhadi, A., Hajishirzi, H., and Smith,
  N.
\newblock Fine-tuning pretrained language models: Weight initializations, data
  orders, and early stopping.
\newblock \emph{arXiv preprint arXiv:2002.06305}, 2020.

\bibitem[Dong et~al.(2022)Dong, Li, Dai, Zheng, Wu, Chang, Sun, Xu, and
  Sui]{dong2022survey}
Dong, Q., Li, L., Dai, D., Zheng, C., Wu, Z., Chang, B., Sun, X., Xu, J., and
  Sui, Z.
\newblock A survey for in-context learning.
\newblock \emph{arXiv preprint arXiv:2301.00234}, 2022.

\bibitem[Du et~al.(2023)Du, Li, Torralba, Tenenbaum, and
  Mordatch]{du2023improving}
Du, Y., Li, S., Torralba, A., Tenenbaum, J.~B., and Mordatch, I.
\newblock Improving factuality and reasoning in language models through
  multiagent debate, 2023.

\bibitem[Eldan \& Russinovich(2023)Eldan and Russinovich]{eldan2023s}
Eldan, R. and Russinovich, M.
\newblock Who's harry potter? approximate unlearning in llms.
\newblock \emph{arXiv preprint arXiv:2310.02238}, 2023.

\bibitem[Fudenberg et~al.(1998)Fudenberg, Drew, Levine, and
  Levine]{fudenberg1998theory}
Fudenberg, D., Drew, F., Levine, D.~K., and Levine, D.~K.
\newblock \emph{The theory of learning in games}, volume~2.
\newblock MIT press, 1998.

\bibitem[Ganguli et~al.(2022)Ganguli, Lovitt, Kernion, Askell, Bai, Kadavath,
  Mann, Perez, Schiefer, Ndousse, et~al.]{ganguli2022red}
Ganguli, D., Lovitt, L., Kernion, J., Askell, A., Bai, Y., Kadavath, S., Mann,
  B., Perez, E., Schiefer, N., Ndousse, K., et~al.
\newblock Red teaming language models to reduce harms: Methods, scaling
  behaviors, and lessons learned.
\newblock \emph{arXiv preprint arXiv:2209.07858}, 2022.

\bibitem[Grill et~al.(2020)Grill, Altch{\'e}, Tang, Hubert, Valko, Antonoglou,
  and Munos]{grill2020monte}
Grill, J.-B., Altch{\'e}, F., Tang, Y., Hubert, T., Valko, M., Antonoglou, I.,
  and Munos, R.
\newblock Monte-carlo tree search as regularized policy optimization.
\newblock In \emph{International Conference on Machine Learning}, pp.\
  3769--3778. PMLR, 2020.

\bibitem[Guo et~al.(2024)Guo, Yao, Shen, Wei, Zhang, Wang, and
  Liu]{guo2024humaninstructionfree}
Guo, H., Yao, Y., Shen, W., Wei, J., Zhang, X., Wang, Z., and Liu, Y.
\newblock Human-instruction-free llm self-alignment with limited samples, 2024.

\bibitem[Guo et~al.(2021)Guo, Wu, Huang, and Xing]{guo2021adversarial}
Guo, W., Wu, X., Huang, S., and Xing, X.
\newblock Adversarial policy learning in two-player competitive games.
\newblock In \emph{International Conference on Machine Learning}, pp.\
  3910--3919. PMLR, 2021.

\bibitem[Guo et~al.(2014)Guo, Singh, Lee, Lewis, and Wang]{guo2014deep}
Guo, X., Singh, S., Lee, H., Lewis, R.~L., and Wang, X.
\newblock Deep learning for real-time atari game play using offline monte-carlo
  tree search planning.
\newblock \emph{Advances in neural information processing systems}, 27, 2014.

\bibitem[Gupta et~al.(2018)Gupta, Mendonca, Liu, Abbeel, and
  Levine]{gupta2018meta}
Gupta, A., Mendonca, R., Liu, Y., Abbeel, P., and Levine, S.
\newblock Meta-reinforcement learning of structured exploration strategies.
\newblock \emph{Advances in neural information processing systems}, 31, 2018.

\bibitem[Gupta et~al.(2019)Gupta, Kumar, Lynch, Levine, and
  Hausman]{gupta2019relay}
Gupta, A., Kumar, V., Lynch, C., Levine, S., and Hausman, K.
\newblock Relay policy learning: Solving long-horizon tasks via imitation and
  reinforcement learning, 2019.

\bibitem[Ha \& Schmidhuber(2018)Ha and Schmidhuber]{ha2018world}
Ha, D. and Schmidhuber, J.
\newblock World models.
\newblock \emph{arXiv preprint arXiv:1803.10122}, 2018.

\bibitem[Hao et~al.(2023)Hao, Gu, Ma, Hong, Wang, Wang, and
  Hu]{hao2023reasoning}
Hao, S., Gu, Y., Ma, H., Hong, J.~J., Wang, Z., Wang, D.~Z., and Hu, Z.
\newblock Reasoning with language model is planning with world model.
\newblock \emph{arXiv preprint arXiv:2305.14992}, 2023.

\bibitem[Heinrich(2017)]{heinrich2017reinforcement}
Heinrich, J.
\newblock \emph{Reinforcement learning from self-play in imperfect-information
  games}.
\newblock PhD thesis, UCL (University College London), 2017.

\bibitem[Hu \& Wellman(2003)Hu and Wellman]{hu2003nash}
Hu, J. and Wellman, M.~P.
\newblock Nash q-learning for general-sum stochastic games.
\newblock \emph{Journal of machine learning research}, 4\penalty0
  (Nov):\penalty0 1039--1069, 2003.

\bibitem[Janner et~al.(2021)Janner, Li, and Levine]{janner2021offline}
Janner, M., Li, Q., and Levine, S.
\newblock Offline reinforcement learning as one big sequence modeling problem.
\newblock \emph{Advances in neural information processing systems},
  34:\penalty0 1273--1286, 2021.

\bibitem[Ji et~al.(2023)Ji, Lee, Frieske, Yu, Su, Xu, Ishii, Bang, Madotto, and
  Fung]{ji2023survey}
Ji, Z., Lee, N., Frieske, R., Yu, T., Su, D., Xu, Y., Ishii, E., Bang, Y.~J.,
  Madotto, A., and Fung, P.
\newblock Survey of hallucination in natural language generation.
\newblock \emph{ACM Computing Surveys}, 55\penalty0 (12):\penalty0 1--38, 2023.

\bibitem[Kahn et~al.(2017)Kahn, Villaflor, Pong, Abbeel, and
  Levine]{kahn2017uncertainty}
Kahn, G., Villaflor, A., Pong, V., Abbeel, P., and Levine, S.
\newblock Uncertainty-aware reinforcement learning for collision avoidance.
\newblock \emph{arXiv preprint arXiv:1702.01182}, 2017.

\bibitem[Kojima et~al.(2022)Kojima, Gu, Reid, Matsuo, and
  Iwasawa]{kojima2022large}
Kojima, T., Gu, S.~S., Reid, M., Matsuo, Y., and Iwasawa, Y.
\newblock Large language models are zero-shot reasoners.
\newblock \emph{Advances in neural information processing systems},
  35:\penalty0 22199--22213, 2022.

\bibitem[Kojima et~al.(2023)Kojima, Gu, Reid, Matsuo, and
  Iwasawa]{kojima2023large}
Kojima, T., Gu, S.~S., Reid, M., Matsuo, Y., and Iwasawa, Y.
\newblock Large language models are zero-shot reasoners, 2023.

\bibitem[Kowalski \& Miernik(2023)Kowalski and
  Miernik]{kowalski2023summarizing}
Kowalski, J. and Miernik, R.
\newblock Summarizing strategy card game ai competition.
\newblock \emph{arXiv preprint arXiv:2305.11814}, 2023.

\bibitem[Lampinen et~al.(2022)Lampinen, Dasgupta, Chan, Matthewson, Tessler,
  Creswell, McClelland, Wang, and Hill]{lampinen2022can}
Lampinen, A.~K., Dasgupta, I., Chan, S.~C., Matthewson, K., Tessler, M.~H.,
  Creswell, A., McClelland, J.~L., Wang, J.~X., and Hill, F.
\newblock Can language models learn from explanations in context?
\newblock \emph{arXiv preprint arXiv:2204.02329}, 2022.

\bibitem[Lanctot et~al.(2017)Lanctot, Zambaldi, Gruslys, Lazaridou, Tuyls,
  P{\'e}rolat, Silver, and Graepel]{lanctot2017unified}
Lanctot, M., Zambaldi, V., Gruslys, A., Lazaridou, A., Tuyls, K., P{\'e}rolat,
  J., Silver, D., and Graepel, T.
\newblock A unified game-theoretic approach to multiagent reinforcement
  learning.
\newblock \emph{Advances in neural information processing systems}, 30, 2017.

\bibitem[LeCun(2022)]{lecun2022path}
LeCun, Y.
\newblock A path towards autonomous machine intelligence version 0.9. 2,
  2022-06-27.
\newblock 2022.

\bibitem[Lee et~al.(2023)Lee, Phatale, Mansoor, Lu, Mesnard, Bishop, Carbune,
  and Rastogi]{lee2023rlaif}
Lee, H., Phatale, S., Mansoor, H., Lu, K., Mesnard, T., Bishop, C., Carbune,
  V., and Rastogi, A.
\newblock Rlaif: Scaling reinforcement learning from human feedback with ai
  feedback.
\newblock \emph{arXiv preprint arXiv:2309.00267}, 2023.

\bibitem[Li et~al.(2023)Li, Cheng, Zhao, Nie, and Wen]{li2023halueval}
Li, J., Cheng, X., Zhao, W.~X., Nie, J.-Y., and Wen, J.-R.
\newblock Halueval: A large-scale hallucination evaluation benchmark for large
  language models.
\newblock In \emph{EMNLP}, pp.\  6449--6464, 2023.

\bibitem[Littman(1994)]{littman1994markov}
Littman, M.~L.
\newblock Markov games as a framework for multi-agent reinforcement learning.
\newblock In \emph{Machine learning proceedings 1994}, pp.\  157--163.
  Elsevier, 1994.

\bibitem[Liu et~al.(2022)Liu, Tam, Muqeeth, Mohta, Huang, Bansal, and
  Raffel]{liu2022few}
Liu, H., Tam, D., Muqeeth, M., Mohta, J., Huang, T., Bansal, M., and Raffel,
  C.~A.
\newblock Few-shot parameter-efficient fine-tuning is better and cheaper than
  in-context learning.
\newblock \emph{Advances in Neural Information Processing Systems},
  35:\penalty0 1950--1965, 2022.

\bibitem[Liu et~al.(2023)Liu, Yao, Ton, Zhang, Cheng, Klochkov, Taufiq, and
  Li]{liu2023trustworthy}
Liu, Y., Yao, Y., Ton, J.-F., Zhang, X., Cheng, R. G.~H., Klochkov, Y., Taufiq,
  M.~F., and Li, H.
\newblock Trustworthy llms: a survey and guideline for evaluating large
  language models' alignment.
\newblock \emph{arXiv preprint arXiv:2308.05374}, 2023.

\bibitem[Min et~al.(2021)Min, Lewis, Zettlemoyer, and
  Hajishirzi]{min2021metaicl}
Min, S., Lewis, M., Zettlemoyer, L., and Hajishirzi, H.
\newblock Metaicl: Learning to learn in context.
\newblock \emph{arXiv preprint arXiv:2110.15943}, 2021.

\bibitem[Min et~al.(2022)Min, Lyu, Holtzman, Artetxe, Lewis, Hajishirzi, and
  Zettlemoyer]{min2022rethinking}
Min, S., Lyu, X., Holtzman, A., Artetxe, M., Lewis, M., Hajishirzi, H., and
  Zettlemoyer, L.
\newblock Rethinking the role of demonstrations: What makes in-context learning
  work?
\newblock \emph{arXiv preprint arXiv:2202.12837}, 2022.

\bibitem[Mordatch(2017)]{learning-to-communicate}
Mordatch, I.
\newblock Learning to communicate, 2017.
\newblock [Online; accessed 18-Jan-2024].

\bibitem[Mordatch \& Abbeel(2018)Mordatch and Abbeel]{mordatch2018emergence}
Mordatch, I. and Abbeel, P.
\newblock Emergence of grounded compositional language in multi-agent
  populations.
\newblock In \emph{Proceedings of the AAAI conference on artificial
  intelligence}, volume~32, 2018.

\bibitem[Nagabandi et~al.(2018)Nagabandi, Clavera, Liu, Fearing, Abbeel,
  Levine, and Finn]{nagabandi2018learning}
Nagabandi, A., Clavera, I., Liu, S., Fearing, R.~S., Abbeel, P., Levine, S.,
  and Finn, C.
\newblock Learning to adapt in dynamic, real-world environments through
  meta-reinforcement learning.
\newblock \emph{arXiv preprint arXiv:1803.11347}, 2018.

\bibitem[Nash(1953)]{nash1953two}
Nash, J.
\newblock Two-person cooperative games.
\newblock \emph{Econometrica: Journal of the Econometric Society}, pp.\
  128--140, 1953.

\bibitem[OpenAI(2023)]{openai2023gpt4}
OpenAI.
\newblock Gpt-4 technical report, 2023.

\bibitem[Ouyang et~al.(2022)Ouyang, Wu, Jiang, Almeida, Wainwright, Mishkin,
  Zhang, Agarwal, Slama, Ray, et~al.]{ouyang2022training}
Ouyang, L., Wu, J., Jiang, X., Almeida, D., Wainwright, C., Mishkin, P., Zhang,
  C., Agarwal, S., Slama, K., Ray, A., et~al.
\newblock Training language models to follow instructions with human feedback.
\newblock In \emph{Proceedings of NeurIPS}, 2022.
\newblock URL \url{https://arxiv.org/abs/2203.02155}.

\bibitem[Perez et~al.(2022)Perez, Huang, Song, Cai, Ring, Aslanides, Glaese,
  McAleese, and Irving]{perez2022red}
Perez, E., Huang, S., Song, F., Cai, T., Ring, R., Aslanides, J., Glaese, A.,
  McAleese, N., and Irving, G.
\newblock Red teaming language models with language models.
\newblock \emph{arXiv preprint arXiv:2202.03286}, 2022.

\bibitem[Perolat et~al.(2021)Perolat, Munos, Lespiau, Omidshafiei, Rowland,
  Ortega, Burch, Anthony, Balduzzi, De~Vylder, et~al.]{perolat2021poincare}
Perolat, J., Munos, R., Lespiau, J.-B., Omidshafiei, S., Rowland, M., Ortega,
  P., Burch, N., Anthony, T., Balduzzi, D., De~Vylder, B., et~al.
\newblock From poincar{\'e} recurrence to convergence in imperfect information
  games: Finding equilibrium via regularization.
\newblock In \emph{International Conference on Machine Learning}, pp.\
  8525--8535. PMLR, 2021.

\bibitem[Pinker(2007)]{pinker2007stuff}
Pinker, S.
\newblock \emph{The stuff of thought: Language as a window into human nature}.
\newblock Penguin, 2007.

\bibitem[Radford et~al.(2019)Radford, Wu, Child, Luan, Amodei, Sutskever,
  et~al.]{radford2019language}
Radford, A., Wu, J., Child, R., Luan, D., Amodei, D., Sutskever, I., et~al.
\newblock Language models are unsupervised multitask learners.
\newblock \emph{OpenAI blog}, 1\penalty0 (8):\penalty0 9, 2019.

\bibitem[Rafailov et~al.(2023)Rafailov, Sharma, Mitchell, Ermon, Manning, and
  Finn]{rafailov2023direct}
Rafailov, R., Sharma, A., Mitchell, E., Ermon, S., Manning, C.~D., and Finn, C.
\newblock Direct preference optimization: Your language model is secretly a
  reward model.
\newblock \emph{arXiv preprint arXiv:2305.18290}, 2023.

\bibitem[Rakelly et~al.(2019)Rakelly, Zhou, Finn, Levine, and
  Quillen]{rakelly2019efficient}
Rakelly, K., Zhou, A., Finn, C., Levine, S., and Quillen, D.
\newblock Efficient off-policy meta-reinforcement learning via probabilistic
  context variables.
\newblock In \emph{International conference on machine learning}, pp.\
  5331--5340. PMLR, 2019.

\bibitem[Saunders et~al.(2022)Saunders, Yeh, Wu, Bills, Ouyang, Ward, and
  Leike]{saunders2022self}
Saunders, W., Yeh, C., Wu, J., Bills, S., Ouyang, L., Ward, J., and Leike, J.
\newblock Self-critiquing models for assisting human evaluators.
\newblock \emph{arXiv preprint arXiv:2206.05802}, 2022.

\bibitem[Schmid(2021)]{schmid2021search}
Schmid, M.
\newblock \emph{Search in imperfect information games}.
\newblock PhD thesis, Charles University, 2021.

\bibitem[Schrittwieser et~al.(2020)Schrittwieser, Antonoglou, Hubert, Simonyan,
  Sifre, Schmitt, Guez, Lockhart, Hassabis, Graepel,
  et~al.]{schrittwieser2020mastering}
Schrittwieser, J., Antonoglou, I., Hubert, T., Simonyan, K., Sifre, L.,
  Schmitt, S., Guez, A., Lockhart, E., Hassabis, D., Graepel, T., et~al.
\newblock Mastering atari, go, chess and shogi by planning with a learned
  model.
\newblock \emph{Nature}, 588\penalty0 (7839):\penalty0 604--609, 2020.

\bibitem[Schulman et~al.(2017)Schulman, Wolski, Dhariwal, Radford, and
  Klimov]{schulman2017proximal}
Schulman, J., Wolski, F., Dhariwal, P., Radford, A., and Klimov, O.
\newblock Proximal policy optimization algorithms.
\newblock \emph{arXiv preprint arXiv:1707.06347}, 2017.

\bibitem[Shoham \& Leyton-Brown(2008)Shoham and
  Leyton-Brown]{shoham2008multiagent}
Shoham, Y. and Leyton-Brown, K.
\newblock \emph{Multiagent systems: Algorithmic, game-theoretic, and logical
  foundations}.
\newblock Cambridge University Press, 2008.

\bibitem[Silver et~al.(2017{\natexlab{a}})Silver, Hubert, Schrittwieser,
  Antonoglou, Lai, Guez, Lanctot, Sifre, Kumaran, Graepel,
  et~al.]{silver2017mastering}
Silver, D., Hubert, T., Schrittwieser, J., Antonoglou, I., Lai, M., Guez, A.,
  Lanctot, M., Sifre, L., Kumaran, D., Graepel, T., et~al.
\newblock Mastering chess and shogi by self-play with a general reinforcement
  learning algorithm.
\newblock \emph{arXiv preprint arXiv:1712.01815}, 2017{\natexlab{a}}.

\bibitem[Silver et~al.(2017{\natexlab{b}})Silver, Schrittwieser, Simonyan,
  Antonoglou, Huang, Guez, Hubert, Baker, Lai, Bolton,
  et~al.]{silver2017alphazero}
Silver, D., Schrittwieser, J., Simonyan, K., Antonoglou, I., Huang, A., Guez,
  A., Hubert, T., Baker, L., Lai, M., Bolton, A., et~al.
\newblock Mastering the game of go without human knowledge.
\newblock \emph{nature}, 550\penalty0 (7676):\penalty0 354--359,
  2017{\natexlab{b}}.

\bibitem[Sokota et~al.(2022)Sokota, D'Orazio, Kolter, Loizou, Lanctot,
  Mitliagkas, Brown, and Kroer]{sokota2022unified}
Sokota, S., D'Orazio, R., Kolter, J.~Z., Loizou, N., Lanctot, M., Mitliagkas,
  I., Brown, N., and Kroer, C.
\newblock A unified approach to reinforcement learning, quantal response
  equilibria, and two-player zero-sum games.
\newblock In \emph{The Eleventh International Conference on Learning
  Representations}, 2022.

\bibitem[Sun et~al.(2023)Sun, Shen, Zhou, Zhang, Chen, Cox, Yang, and
  Gan]{sun2023principle}
Sun, Z., Shen, Y., Zhou, Q., Zhang, H., Chen, Z., Cox, D., Yang, Y., and Gan,
  C.
\newblock Principle-driven self-alignment of language models from scratch with
  minimal human supervision.
\newblock \emph{arXiv preprint arXiv:2305.03047}, 2023.

\bibitem[Sutton et~al.(1999)Sutton, McAllester, Singh, and
  Mansour]{sutton1999policy}
Sutton, R.~S., McAllester, D., Singh, S., and Mansour, Y.
\newblock Policy gradient methods for reinforcement learning with function
  approximation.
\newblock \emph{Advances in neural information processing systems}, 12, 1999.

\bibitem[Team(2023)]{geminiteam2023gemini}
Team, G.
\newblock Gemini: A family of highly capable multimodal models, 2023.

\bibitem[Tesauro(1995)]{tesauro1995td}
Tesauro, G.
\newblock Td-gammon: A self-teaching backgammon program.
\newblock In \emph{Applications of neural networks}, pp.\  267--285. Springer,
  1995.

\bibitem[Touvron et~al.(2023)Touvron, Lavril, Izacard, Martinet, Lachaux,
  Lacroix, Rozi{\`e}re, Goyal, Hambro, Azhar, et~al.]{touvron2023llama}
Touvron, H., Lavril, T., Izacard, G., Martinet, X., Lachaux, M.-A., Lacroix,
  T., Rozi{\`e}re, B., Goyal, N., Hambro, E., Azhar, F., et~al.
\newblock Llama: Open and efficient foundation language models.
\newblock \emph{arXiv preprint arXiv:2302.13971}, 2023.

\bibitem[Vaswani et~al.(2017)Vaswani, Shazeer, Parmar, Uszkoreit, Jones, Gomez,
  Kaiser, and Polosukhin]{vaswani2017attention}
Vaswani, A., Shazeer, N., Parmar, N., Uszkoreit, J., Jones, L., Gomez, A.~N.,
  Kaiser, {\L}., and Polosukhin, I.
\newblock Attention is all you need.
\newblock \emph{Advances in neural information processing systems}, 30, 2017.

\bibitem[Vinyals et~al.(2019)Vinyals, Babuschkin, Czarnecki, Mathieu, Dudzik,
  Chung, Choi, Powell, Ewalds, Georgiev, et~al.]{vinyals2019grandmaster}
Vinyals, O., Babuschkin, I., Czarnecki, W.~M., Mathieu, M., Dudzik, A., Chung,
  J., Choi, D.~H., Powell, R., Ewalds, T., Georgiev, P., et~al.
\newblock Grandmaster level in starcraft ii using multi-agent reinforcement
  learning.
\newblock \emph{Nature}, 575\penalty0 (7782):\penalty0 350--354, 2019.

\bibitem[Wang et~al.(2023)Wang, Chen, Pei, Xie, Kang, Zhang, Xu, Xiong, Dutta,
  Schaeffer, et~al.]{wang2023decodingtrust}
Wang, B., Chen, W., Pei, H., Xie, C., Kang, M., Zhang, C., Xu, C., Xiong, Z.,
  Dutta, R., Schaeffer, R., et~al.
\newblock Decodingtrust: A comprehensive assessment of trustworthiness in gpt
  models.
\newblock \emph{arXiv preprint arXiv:2306.11698}, 2023.

\bibitem[Wang et~al.(2020)Wang, Liu, and Li]{wang2020reinforcement}
Wang, J., Liu, Y., and Li, B.
\newblock Reinforcement learning with perturbed rewards.
\newblock In \emph{Proceedings of the AAAI conference on artificial
  intelligence}, volume~34, pp.\  6202--6209, 2020.

\bibitem[Wang et~al.(2021)Wang, Guo, Zhu, and Liu]{wang2021policy}
Wang, J., Guo, H., Zhu, Z., and Liu, Y.
\newblock Policy learning using weak supervision.
\newblock \emph{Advances in Neural Information Processing Systems},
  34:\penalty0 19960--19973, 2021.

\bibitem[Wang et~al.(2022{\natexlab{a}})Wang, Wei, Schuurmans, Le, Chi, Narang,
  Chowdhery, and Zhou]{wang2022selfcot}
Wang, X., Wei, J., Schuurmans, D., Le, Q., Chi, E., Narang, S., Chowdhery, A.,
  and Zhou, D.
\newblock Self-consistency improves chain of thought reasoning in language
  models.
\newblock \emph{arXiv preprint arXiv:2203.11171}, 2022{\natexlab{a}}.

\bibitem[Wang et~al.(2022{\natexlab{b}})Wang, Kordi, Mishra, Liu, Smith,
  Khashabi, and Hajishirzi]{wang2022self}
Wang, Y., Kordi, Y., Mishra, S., Liu, A., Smith, N.~A., Khashabi, D., and
  Hajishirzi, H.
\newblock Self-instruct: Aligning language model with self generated
  instructions.
\newblock \emph{arXiv preprint arXiv:2212.10560}, 2022{\natexlab{b}}.

\bibitem[Wei et~al.(2022)Wei, Wang, Schuurmans, Bosma, Xia, Chi, Le, Zhou,
  et~al.]{wei2022chain}
Wei, J., Wang, X., Schuurmans, D., Bosma, M., Xia, F., Chi, E., Le, Q.~V.,
  Zhou, D., et~al.
\newblock Chain-of-thought prompting elicits reasoning in large language
  models.
\newblock \emph{Advances in Neural Information Processing Systems},
  35:\penalty0 24824--24837, 2022.

\bibitem[Wen et~al.(2022)Wen, Kuba, Lin, Zhang, Wen, Wang, and
  Yang]{wen2022multi}
Wen, M., Kuba, J., Lin, R., Zhang, W., Wen, Y., Wang, J., and Yang, Y.
\newblock Multi-agent reinforcement learning is a sequence modeling problem.
\newblock \emph{Advances in Neural Information Processing Systems},
  35:\penalty0 16509--16521, 2022.

\bibitem[White et~al.(2023)White, Fu, Hays, Sandborn, Olea, Gilbert, Elnashar,
  Spencer-Smith, and Schmidt]{white2023prompt}
White, J., Fu, Q., Hays, S., Sandborn, M., Olea, C., Gilbert, H., Elnashar, A.,
  Spencer-Smith, J., and Schmidt, D.~C.
\newblock A prompt pattern catalog to enhance prompt engineering with chatgpt.
\newblock \emph{arXiv preprint arXiv:2302.11382}, 2023.

\bibitem[Xi et~al.(2023)Xi, Zhang, Xiao, Huang, Deng, Liang, Chen, and
  Sun]{xi2023mastering}
Xi, W., Zhang, Y., Xiao, C., Huang, X., Deng, S., Liang, H., Chen, J., and Sun,
  P.
\newblock Mastering strategy card game (legends of code and magic) via
  end-to-end policy and optimistic smooth fictitious play.
\newblock \emph{arXiv preprint arXiv:2303.04096}, 2023.

\bibitem[Xie et~al.(2021)Xie, Raghunathan, Liang, and Ma]{xie2021explanation}
Xie, S.~M., Raghunathan, A., Liang, P., and Ma, T.
\newblock An explanation of in-context learning as implicit bayesian inference.
\newblock \emph{arXiv preprint arXiv:2111.02080}, 2021.

\bibitem[Xu(2016)]{xu2016convergence}
Xu, Z.
\newblock Convergence of best-response dynamics in extensive-form games.
\newblock \emph{Journal of Economic Theory}, 162:\penalty0 21--54, 2016.

\bibitem[Yao et~al.(2023)Yao, Xu, and Liu]{yao2023large}
Yao, Y., Xu, X., and Liu, Y.
\newblock Large language model unlearning.
\newblock \emph{arXiv preprint arXiv:2310.10683}, 2023.

\bibitem[Zhang et~al.(2023{\natexlab{a}})Zhang, Chen, Wang, Xie, Liu, Lui, and
  Li]{zhang2023uncertainty}
Zhang, X., Chen, J., Wang, H., Xie, H., Liu, Y., Lui, J.~C., and Li, H.
\newblock Uncertainty-aware instance reweighting for off-policy learning.
\newblock In \emph{Thirty-seventh Conference on Neural Information Processing
  Systems}, 2023{\natexlab{a}}.

\bibitem[Zhang et~al.(2023{\natexlab{b}})Zhang, Li, Cui, Cai, Liu, Fu, Huang,
  Zhao, Zhang, Chen, et~al.]{zhang2023siren}
Zhang, Y., Li, Y., Cui, L., Cai, D., Liu, L., Fu, T., Huang, X., Zhao, E.,
  Zhang, Y., Chen, Y., et~al.
\newblock Siren's song in the ai ocean: A survey on hallucination in large
  language models.
\newblock \emph{arXiv preprint arXiv:2309.01219}, 2023{\natexlab{b}}.

\bibitem[Zhou et~al.(2022{\natexlab{a}})Zhou, Sch{\"a}rli, Hou, Wei, Scales,
  Wang, Schuurmans, Cui, Bousquet, Le, et~al.]{zhou2022least}
Zhou, D., Sch{\"a}rli, N., Hou, L., Wei, J., Scales, N., Wang, X., Schuurmans,
  D., Cui, C., Bousquet, O., Le, Q., et~al.
\newblock Least-to-most prompting enables complex reasoning in large language
  models.
\newblock \emph{arXiv preprint arXiv:2205.10625}, 2022{\natexlab{a}}.

\bibitem[Zhou et~al.(2023)Zhou, Schärli, Hou, Wei, Scales, Wang, Schuurmans,
  Cui, Bousquet, Le, and Chi]{zhou2023leasttomost}
Zhou, D., Schärli, N., Hou, L., Wei, J., Scales, N., Wang, X., Schuurmans, D.,
  Cui, C., Bousquet, O., Le, Q., and Chi, E.
\newblock Least-to-most prompting enables complex reasoning in large language
  models, 2023.

\bibitem[Zhou et~al.(2022{\natexlab{b}})Zhou, Muresanu, Han, Paster, Pitis,
  Chan, and Ba]{zhou2022large}
Zhou, Y., Muresanu, A.~I., Han, Z., Paster, K., Pitis, S., Chan, H., and Ba, J.
\newblock Large language models are human-level prompt engineers.
\newblock \emph{arXiv preprint arXiv:2211.01910}, 2022{\natexlab{b}}.

\bibitem[Zhu et~al.(2023{\natexlab{a}})Zhu, Lin, Jain, and
  Zhou]{zhu2023transfer}
Zhu, Z., Lin, K., Jain, A.~K., and Zhou, J.
\newblock Transfer learning in deep reinforcement learning: A survey.
\newblock \emph{IEEE Transactions on Pattern Analysis and Machine
  Intelligence}, 2023{\natexlab{a}}.

\bibitem[Zhu et~al.(2023{\natexlab{b}})Zhu, Wang, Cheng, and
  Liu]{zhu2023unmasking}
Zhu, Z., Wang, J., Cheng, H., and Liu, Y.
\newblock Unmasking and improving data credibility: A study with datasets for
  training harmless language models.
\newblock \emph{arXiv preprint arXiv:2311.11202}, 2023{\natexlab{b}}.

\bibitem[Ziegler et~al.(2019)Ziegler, Stiennon, Wu, Brown, Radford, Amodei,
  Christiano, and Irving]{ziegler2019fine}
Ziegler, D.~M., Stiennon, N., Wu, J., Brown, T.~B., Radford, A., Amodei, D.,
  Christiano, P., and Irving, G.
\newblock Fine-tuning language models from human preferences.
\newblock \emph{arXiv preprint arXiv:1909.08593}, 2019.

\end{thebibliography}

\end{document}